# AUGMENTED CROSS-SELLING THROUGH EXPLAINABLE AI—A CASE FROM ENERGY RETAILING

*Research Paper*

Felix Haag, University of Bamberg, Bamberg, Germany, felix.haag@uni-bamberg.de

Konstantin Hopf, University of Bamberg, Bamberg, Germany, konstantin.hopf@uni-bamberg.de

Pedro Menelau Vasconcelos, University of Bamberg, Bamberg, Germany, pedro.menelau@icloud.com

Thorsten Staake, University of Bamberg, Bamberg, Germany, thorsten.staake@uni-bamberg.de; ETH Zurich, Zurich, Switzerland

## Abstract

*The advance of Machine Learning (ML) has led to a strong interest in this technology to support decision making. While complex ML models provide predictions that are often more accurate than those of traditional tools, such models often hide the reasoning behind the prediction from their users, which can lead to lower adoption and lack of insight. Motivated by this tension, research has put forth Explainable Artificial Intelligence (XAI) techniques that uncover patterns discovered by ML. Despite the high hopes in both ML and XAI, there is little empirical evidence of the benefits to traditional businesses. To this end, we analyze data on 220,185 customers of an energy retailer, predict cross-purchases with up to 86% correctness (AUC), and show that the XAI method SHAP provides explanations that hold for actual buyers. We further outline implications for research in information systems, XAI, and relationship marketing.*

## 1 Motivation

Relationship marketing aims to systematically build long-term relationships with customers. As the attraction of new customers is costly, relationship marketing focuses on developing and "nurturing" existing customers (Kumar, 2018). Activities include customer loyalty programs, the selling of higher-valued offerings (up-selling) or additional products (cross-selling). When done right, cross-selling promises significant profits for companies (Kamakura, 2008; Schmitz et al., 2014). To support this practice, organizations invest into customer relationship management (CRM) systems, which systematically record customer data (e.g., order and payment history) and support a wide variety of organizational tasks in relationship marketing (Zablah et al., 2012). Yet, implementing a CRM system does not always lead to the desired business impact (Chang et al., 2014; Zablah et al., 2012). Also, organisations face challenges in actually realising business value from big data in CRM systems (Sivarajah et al., 2017) and lack knowledge on how to condense the rich information to operational insights (Kitchens et al., 2018).

Machine learning (ML), the core technology of artificial intelligence (AI) for recognizing patterns in data and making predictions, has recently made strong advances in improving modelling capabilities. This also led to new applications for relationship marketing: Several studies have already demonstrated, for example, how to derive customer spending in promotional campaigns (Shrivastava and Jank, 2015), purchase probabilities (Martens et al., 2016), customer profitability (Cui et al., 2012), and customer characteristics (Hopf et al., 2016). Such fine-grained insights represent a significant contribution to operational sales processes in cross-selling. Despite illustrative examples and the rapid development of ML, it seems that firms rarely achieve productivity gains from respective investments (Müller et al., 2018; Tambe, 2014; Wu et al., 2019)—Brynjolfsson et al. (2017) also refer to this problem as the AI productivity paradox.

The benefits of ML technology are particularly hard to realize in areas where processes have a high human involvement, such as retail marketing. Processes that involve humans require technologies that are geared towards users, i.e., they should "augment" human work (Grønsund and Aanestad, 2020). However, only a few empirical examples of ML-based work augmentation exist in practice. Research on ML in relationship marketing so far focuses strongly on targeting (i.e., selecting most promising customers) and recommender systems studies have concentrated on automated tailoring (e.g., finding the right product to offer) (Wang and Benbasat, 2016; Wang and Wang, 2019), but not on how technologies can augment the work of sales agents that seek to cross-sell products.

Energy utilities belong to a sector with established customer relationships due to its contract-based offerings. They have increasingly comprehensive customer data due to ongoing digitization and the deployment of smart meter infrastructures. Formerly often monopolists, these incumbents now frequently operate in liberalized, competitive markets (Cramton, 2017). Energy utilities have therefore a growing interest in new ways to maintain profits through cross-selling efforts. As only few studies investigate the application of ML in the area of cross-selling, we investigate the following research question (RQ):

> *RQ1: To what extent can ML predict future cross-buying behavior in energy retailing based on existing purchase behavior data?*

Answering RQ1 demonstrates ability of ML to aid cross-selling processes and helps especially traditional firms with their decision on investing into this technology. For research, it lays the foundation to further investigate issues around the organizational use of ML (Baier et al., 2019; Berente et al., 2021). One of the issues is that most well-functioning and contemporary ML applications are "black boxes" (Guidotti et al., 2018). Thus, they are inscrutable and difficult for humans to understand (Berente et al., 2021). This causes several problems that hinder the broader adoption of ML applications in corporate systems, like those used in CRM. First, the inscrutability brings user acceptance problems (Burton et al., 2020; Dietvorst et al., 2018). Second, for many applications, predictions alone are only of limited help and users expect prescriptions with concrete recommendations for action (Mehdiyev and Fettke, 2020). Third, users want to exploit the pattern recognition capability of the algorithms. They want to see and understand the ML-detected patterns in the data instead of receiving predictions (Berente et al., 2019; Tremblay et al., 2021).

Research on knowledge-based systems has identified explanations as a potential remedy to some of these problems (Gregor and Benbasat, 1999; Wang and Benbasat, 2007) and ML research has put forth Explainable Artificial Intelligence (XAI) techniques. XAI methods provide many opportunities to offer additional insights for relationship marketing, but current approaches have a strong focus on technical aspects or data perspectives of developers. Thus, the outputs of these approaches cannot directly be used in information systems (IS) to support humans (Abdul et al., 2018; Miller, 2019; Wastensteiner et al., 2021). Current IS research lacks evaluations regarding the validity and robustness of patterns detected by XAI in the data. Also, the design of explanations for business users remains underinvestigated (Cheng et al., 2019).

For the case of cross-selling, predictions on which products a certain customer is likely to buy—an ML model can predict this information based on behavioral data (Martínez et al., 2020)—is just of limited help, because sales experts need additional insights on arguments that matter for the customer to buy a specific product. Thus, we explore the benefits of XAI methods in this case with:

> *RQ2: How well can XAI derive patterns from predictive models on the example of cross-buying behaviour that hold for actual buyers?*

In answering RQ2, we assess how valid the generated explanations are and provide a robustness check of the obtained insights. Our resulting ML and XAI models lay the foundation for further investigations involving human participants, who can use the insights and generate business value that future studies can quantify.

## 2 Related Work

We briefly review existing approaches of IS research and practice to support sales activities. By catching up on the latest developments in the area of ML applications in that field, we point to the limitations of black box models and applications that support humans in the relationship marketing process.

### 2.1 IS for sales support

IS have a long tradition in supporting the sales processes in companies. This started with systems that organized existing knowledge on customers and products in the form of databases (e.g., Binbasioglu and Jarke, 1986; King, 1978), enterprise reporting and business intelligence systems that aggregate data to descriptive figures (Shollo and Galliers, 2016), and CRM systems that aim to record all interactions of a firm with its customers and to analyze these interactions in order to optimize profitability, customer satisfaction, and customer retention. From a management perspective, CRM systems are often mentioned in the same vein as ERP systems and have a high strategic relevance (Luftman et al., 2012). Literature distinguishes between support-related and targeting-related types of CRM systems (Kim and Mukhopadhyay, 2010). As a direct support for front-line employees, support-related systems (also known as "front-office" or "operational" CRM) store and manage data for providing customized service. Targeting-related systems (often called "analytical," "strategic," or "back-office" CRM) analyze customers' preferences and purchasing behaviors (Kim and Mukhopadhyay, 2010).

Studies in IS have investigated different aspects of CRM systems in an organizational context. One stream examines effective CRM systems design and implementation (Gefen and Ridings, 2002; Kim and Mukhopadhyay, 2010; Ward et al., 2005). This research is supported by studies on the user satisfaction of front-line employees with CRM that leads to a better perceived service quality by customers (Hsieh et al., 2012). Another stream focuses on the influence of CRM systems on firm performance (Coltman, 2007; Coltman et al., 2011; Karimi et al., 2001; Zablah et al., 2012). A third stream concentrates on how customer-centric websites must be built to satisfy customers and improve customer relationships (Albert and Goes, 2004; Lee et al., 2003).

Overall, the primary goal of CRM systems is to store customer data, together with transaction and interaction data. There is growing interest in extracting value from the stored data, as well as connecting and merging open and big data sources to provide additional insights into the customer data (Akter and Fosso Wamba, 2016; Constantiou and Kallinikos, 2015; LaValle et al., 2011).

### 2.2 ML in sales support systems

ML techniques have been used for many years to obtain predictions or prescriptions from data stored in corporate IS (Arnott and Pervan, 2008; Shaw and Tu, 1988). Yet, recent advances in ML algorithms, the availability of data, and computing power allowed extensive use in corporate processes (Akter and Fosso Wamba, 2016; Duan et al., 2019; Tarafdar et al., 2019; Watson, 2017). In contrast to earlier attempts to make IS intelligent, which were primarily based on human-encoded rule sets (Duan et al., 2019; Haenlein and Kaplan, 2019), ML can automatically acquire rules from data and thereby overcome the "knowledge acquisition bottleneck" (Cullen and Bryman, 1988) that has been a major obstacle of rule-based systems. ML-based IS can therefore support the personal selling process of goods at critical steps (Syam and Sharma, 2018). Studies in the field of IS illustrate how ML can already provide guidance in the targeting phase of the personal selling process. For example, studies predicted purchase probabilities for specific products (Loureiro et al., 2018; Martens et al., 2016; Olson and Chae, 2012), customer spending in promotional campaigns (Shrivastava and Jank, 2015), and customer profitability (Cui et al., 2012). However, existing studies on ML in relationship marketing so far focus strongly on targeting (i.e., selecting the most promising customers for a single product), but not on tailoring offers to customers (i.e., finding the right product to offer for a buyer)—a core activity in cross-selling.

## 2.3     Explainable Artifical Intelligence (XAI) in IS

ML methods can obtain powerful predictions, but the resulting models are usually black boxes, which means they are inscrutable to humans (Berente et al., 2021; Guidotti et al., 2018). As explanations are demonstrated to improve user performance (Gregor and Benbasat, 1999), satisfaction (Li and Gregor, 2011), and trust (Yeomans et al., 2019) in intelligent systems, there is a high interest in obtaining explanations for ML models. Explanations can also help to improve the reliability of quantitative models by allowing humans to evaluate the causality of discovered relationships (Rudin, 2019) and match the steps of machine reasoning with existing conceptual or mental models (Tremblay et al., 2021; van den Broek et al., 2021).

Given this interest, XAI has emerged as a very active field of research and has produced various methods to explain the predictions of ML models (Barredo Arrieta et al., 2020). Along with the development of new XAI approaches, three main properties have been mentioned in the literature to distinguish between methods (Adadi and Berrada, 2018; Stiglic et al., 2020): First, the level of interpretability, with *local* methods explaining individual predictions and *global* methods clarifying the entire model behavior that has led to all the individual model outcomes. Second, the applicability of methods, where *model-agnostic* approaches are pluggable to any model, while *specific* methods are limited to a specific class of ML. Third, whether procedures are necessary to increase the interpretability of the model. Explanation methods that explain an already trained model are referred to in the literature as post-hoc methods. One such technique aims to measure the feature's influence on the outcome by permuting feature values for some noise. Intrinsic approaches, in contrast, are considered *interpretable by design,* like linear regression (Adadi and Berrada, 2018; Barredo Arrieta et al., 2020; Molnar, 2019).

We focus on model-agnostic and local methods that explain the attribution of features for a model's outcome. For the cross-selling case, they provide explanations on the level of each customer and identify the features that are most important for a predicted customer behavior. Two feature attribution methods that provide model-agnostic and local explanations are frequently cited in the literature (Slack et al., 2020): First, Local Interpretable Model-agnostic Explanations (LIME), which perturbs data and feeds them into an ML model. Based on this newly generated training data and the model predictions, LIME fits an interpretable model (e.g., linear regression) weighted by the proximity of the synthetically generated observation to the actual observation. Accordingly, the method calculates a feature's attribution through the distance of a prediction of the interpretable model to the non-interpretable model (Ribeiro et al., 2016). Second, Shapley Additive Explanations (SHAP) proposed by Lundberg and Lee (2017) uses concepts from cooperative game theory to determine the extent to which a given feature value affects a single prediction. SHAP aims to assign an impact value to all features at the individual prediction level (i.e., generating local explanations) and thus, belongs to the class of additive feature attribution methods (Molnar, 2019). Shapley and SHAP values in particular aim to uncover how a feature contributes to the change in model prediction for a data instance compared to the average prediction of the model over all data instances (Lundberg et al., 2020). LIME has recently been subject to criticism for providing different explanations for the same instances when repeating explanation generation, resulting in low stability (Carvalho et al., 2019). In fact, LIME randomly perturbs data in the neighborhood of a training instance to obtain inputs for fitting an interpretable model (Ribeiro et al., 2016). SHAP, in contrast, avoids random components to generate explanations for linear and tree-based ML by interpreting the models directly (Lundberg et al., 2020; Lundberg and Lee, 2017). Velmurugan et al. (2021) can show, using several tabular datasets, that SHAP is—as expected—the more stable XAI method for linear and tree-based ML. Also, using time-series data, Schlegel et al. (2019) show that SHAP outperforms LIME in terms of meaningful explanations. Hence, we consider SHAP as the most relevant feature attribution method for the generation of explanations for cross-selling.

Despite the helpful insights that current XAI methods allow into ML models, the methods have a strong focus on technical aspects or the data perspective of developers. The outputs of such methods are therefore not directly usable in IS intended to present information to business users (Abdul et al., 2018; Miller, 2019; Wastensteiner et al., 2021). Rather, research is needed that investigates how explanations should be designed for business users (Cheng et al., 2019).

# 3 Research Approach

To answer our research questions, we conducted an empirical analysis on the use of current ML and XAI methods (Bailey and Barley, 2020; Tremblay et al., 2021) in relationship marketing. In accordance with the guidelines for conducting ML research in IS (Kühl et al., 2021), we give a detailed account of our data analysis in the method and results section.

## 3.1 Case selection, problem description, and study design

From our empirical case study on applying ML in an energy utility's cross-selling efforts, we gained knowledge on the use of AI in organizations, particularly in the relationship marketing context. We purposefully selected (Patton, 2002, p. 234) the energy retailing industry because it is an increasingly important business field. We found a company that provided us with unique data from their operational databases that we can use for scientific inquiry. The investigated application of ML in its richness was—to the best of our knowledge—previously not accessible to IS inquiry, it is thus a "revelatory case study" (Yin, 2018, p. 50), enabling us to study our research question with a single-case study.

The energy utility business has become increasingly competitive in the last years (Cramton, 2017): New companies have entered the market, energy products have become more expensive, and the political debate around counteracting climate change raised new requirements for utility companies. All this is confronted with the fact that the share of the wallet spent on housing, water and energy has remained nearly constant during the last 20 years (Eurostat, 2017). Another problem of energy retailers is that their products are either contractual (like electricity, gas, water, cable TV, Internet, etc.) or durable (like electric vehicle wall boxes, photovoltaic installations, etc.). Purchasing such products requires a relatively high buyer involvement and usually takes place in personal selling situations (Kotler et al., 2017), for example, via a phone call. Consulting customers for such cross-selling products requires significant experience from sales experts (Allcott & Sweeney, 2017; Bhaskaran & Gilbert, 2009). These employees consider the situation of the customer, select one or more suitable product(s) to offer, and argue for the benefits of the product. Many studies from the application of ML in relationship marketing focus on the qualification of customers for a certain product to support targeting. Our case sheds light on the situation where a sales expert stays in contact with an existing customer. In such a situation, it is not only relevant which customer buys a certain product, but also which products are most suitable for the sales expert to offer. Beyond that, it is valuable to know arguments that are most effective in offering the respective products.

Our study followed a typical data analytics approach (Fayyad et al., 1996; Shearer, 2000). We gathered knowledge on the business situation, inspected original data, prepared the data for modeling, applied ML models and an XAI approach, and evaluated the results according to standards of the discipline (Kühl et al., 2021). We illustrate our approach in Figure 1 and describe each step below.

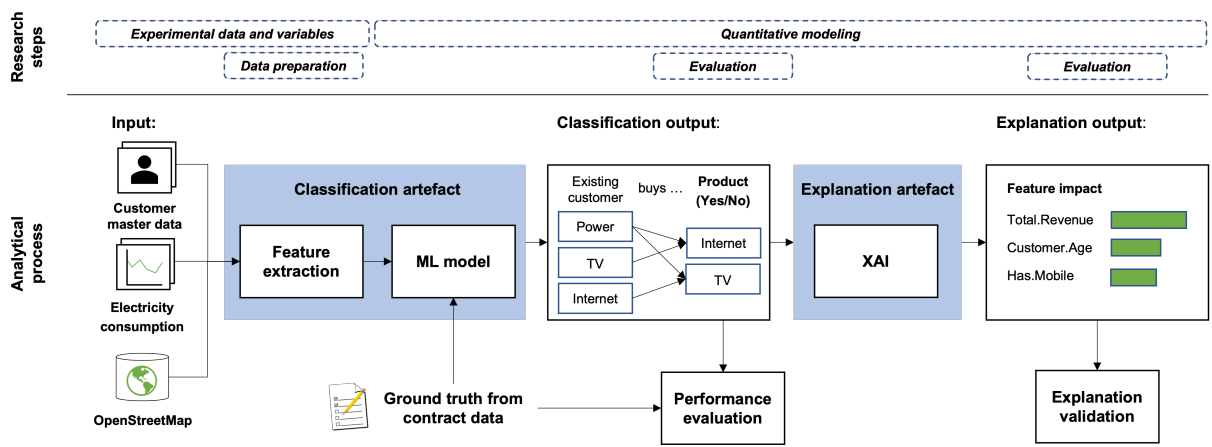

*Figure 1.    Research approach and technical implementation.*

*Augmented Cross-Selling Through Explainable AI*## 3.2 Experimental data and variables

We received a dataset from a European utility company that offers power, gas, Internet, and cable TV contracts to private customers. The dataset consisted of contact events (e.g., emails, phone calls), contract information, yearly electricity consumption for existing contracts, revenue information, and data on the dunnings issued in the case of late payments. In total, the dataset contained 95 variables on 220,185 unique power, Internet, or cable TV customers who were active in the years 2012 through 2017, spanning 547,848 utility service contracts. When formatting the dataset over the timespan to obtain the relevant information for each prediction case and year (e.g., power consumption or revenue for the given year), we came up with a set of 23 unique variables.

To enrich the data with environmental information, we obtained 44 variables from the geographic database OpenStreetMap for each customer address specified in the contracts, following the procedures described by Hopf (2018). These variables describe the geographic area in an area of 300x300m around the customer's location (e.g., distance to businesses, mean area of nearby buildings). Table 1 provides a summary of all variables used for the analysis.

| Group | Variable | Description | Data type |
|---|---|---|---|
| Customer | StartYear | Year of the start of the first contract between customer and company. | Integer |
|  | Age | Age of the customer in years according to the last contract. | Float |
|  | FormOfAddress | Form of address of the customer according to the last contract. | Categorical |
|  | RelationshipMonthsUntil{year} | Number of calendar months since the beginning of the customer relationship until the end of the observed year. | Float |
|  | NumberOfContacts{year} | Number of discrete contact events initiated by the customer. | Float |
|  | BankType | Type of last specified bank. | Categorical |
|  | NumberOfDunnings{year} | Number of dunning notices issued by the company to the customer. | Float |
| Has | Title, Phone, Mobile, Email, Different billing address, IBAN | Binary values stating whether the customer has specified an academic title/landline/mobile phone number/e-mail address/different billing address/international bank account number in any of their contracts or not. | Boolean |
|  | ServicePortal, OnlineBills | Binary values stating if customer uses service portal/online billing option. | Boolean |
| Consumption | NormPower{year} | Yearly average electricity consumption in kWh. | Float |
| Total/Net | Revenue{ContractType}{year} | Net revenue from all cable TV/Internet/power/total contracts of the customer. | Float |
| Existing Customer | Power{year}, Inet{year}, TV{year} | Binary values stating whether the customer is an existing (i.e., customer has had the relevant contract for the entirety of the observed year) or new power, Internet or cable TV customer for a given year or not. | Boolean |
| Purchase (dependent variable) |  |  | Boolean |
| Coordinates | Lat, Long | Latitude, longitude of the customer's main address in degrees. | Float |
| Building | Area | Building area in m² (mean, median, variance). | Float |
|  | NextBuildingArea | Building area of the closest building in m². | Float |
|  | NextBuildingsDist | Distance to the closest building in m (mean, variance). | Float |
|  | BuildingDist | Distance to buildings in m (mean, variance). | Float |
|  | ThisBuildingType, NextBuildingType, BuildingTypeMode | The type of the building, closest building, mode for building type within an area of 300x300m around the customer's location. | Categorical |
| Num | Buildings, PublicInstitutions, Business, Food, Transportation, Recreation, Culture, Sights, Countryside, RoadSystem | Number of buildings/public institutions/businesses/eating places/public transportation/recreational/cultural features/tourist attractions/countryside/roads within an area of 300x300m around the customer's location. | Float |
| MinDist | Business, Food, Culture | Minimum distance to the closest public institution/businesses/eating places/public transportation/recreational/cultural feature/tourist attraction/countryside/road system within an area of 300x300m around the customer's location in m. | Float |
| MeanDist | PublicInstitutions, Business, Food, Transportation, Recreation, Culture, | Mean distance to the closest public institution/businesses/eating places/public transportation/recreational/cultural | Float |

*Thirtieth European Conference on Information Systems (ECIS 2022), Timisoara, Romania*        6

*Augmented Cross-Selling Through Explainable AI*

| Group | Variable | Description | Data type |
|---|---|---|---|
|  | Sights, Countryside, RoadSystem | feature/tourist attraction/countryside/road system within an area of 300x300m around the customer's location in m. |  |
| TotalArea | Apartments, SingleFamily, NonResidential, NotSpecified, CountrySide, Residential, City | Total area for apartment/single family/non-residential/not specified/countryside/residential/city zoning within an area of 300x300m around the customer's location in m$^2$. | Float |
| Land use | ThisLandUseType NextLandUseType | Land use type of the customer's location and of the closest land partition within an area of 300x300m around the customer's location. | Categorical |

*Table 1.  Variables used for cross-selling prediction.*

## 3.3 Data Preparation

In order to prepare the data for analysis, we conducted several preparatory steps. First, we excluded contract data that obviously belonged to business customers. We did so by filtering the salutation (e.g., "Mr." or "Mrs.") and the power contracts with an electricity consumption of less than 100,000 kWh—this was the consumption threshold beyond which the utility company offered specific business contracts. We also excluded all customers without active contracts in the time frame of 2012-2017. Second, we computed the yearly revenue for each TV or Internet contract using each contracts' active months within each year and multiplying them with the monthly price for the respective tariff (e.g., EUR 15.32 for 30 Mbit/s Internet). For power contracts, we calculated the revenue based on the mean daily consumption for each year and the corresponding consumption-oriented rates for each type of contract. Third, we aggregated all contract information up to the customer level by summing up revenues and consumption of all contracts of a customer. When a single customer had contracts under multiple addresses, we used the geographic information with regard to their most frequent address.

Finally, we formatted the dependent variables. If a customer was an existing customer for a type of contract in a given year and cross-purchased a contract of a different type in the following year, the value of the dependent variable *{contractType}Purchase{year}* was *True*, else the value of the dependent variable was *False*. Table 2 shows the distribution of all dependent variables for all years.

| Year | Number of observations | | | |
|---|---|---|---|---|
| (Train/Test) | Power buys Inet | Power buys TV | TV buys Inet | Internet buys TV |
| 2012/2013 | 162,957 (1,508 / 0.9%) | 162,957 (1,565 / 1.0%) | 68,882 (1,159 / 1.7%) | 31,077 (149 / 0.5%) |
| 2013/2014 | 163,885 (1,711 / 1.0%) | 163,885 (2,011 / 1.2%) | 71,115 (1,333 / 1.9%) | 32,902 (200 / 0.6%) |
| 2014/2015 | 164,186 (1,728 / 1.1%) | 164,186 (1,769 / 1.1%) | 72,477 (1,247 / 1.7%) | 34,616 (161 / 0.5%) |
| 2015/2016 | 164,727 (2,087 / 1.3%) | 164,727 (2,964 / 1.8%) | 74,386 (1,322 / 1.8%) | 36,948 (190 / 0.5%) |
| 2016/2017 | 165,553 (1,884 / 1.1%) | 165,553 (2,123 / 1.3%) | 75,461 (1,134 / 1.5%) | 39,278 (263 / 0.7%) |

*Table 2.  Number of observations per train/test year and cross-purchase case (number and ratio of true positives in parentheses).*

## 3.4 Quantitative modeling

Our modeling consisted of three steps. First, we chose the ML algorithms and their specific parameters for the task. Then, we trained the classifiers on each prediction case. Finally, we used the XAI method SHAP to explain the output of the models and validated the explanation output.

The proportion of positive and negative observations displayed a highly imbalanced class structure due to the fact that only a small fraction of the customers represented in the dataset cross-purchased contracts in each year. That is, the percentage of the positive class among all observations in each cross-purchase case and train/test year ranged from 0.5% to 1.9%. Class imbalance can be a particularly difficult obstacle on the path to obtaining useful predictions in ML (Branco et al., 2016). Considering their comparably good performance in a high number of applications (Emanet et al., 2014; Fernandez-Delgado et al., 2014), we decided on applying a variant of the Random Forest (RF) and of the AdaBoost classifier.

Specifically, we used the Balanced RF[1] (Chen et al., 2004) and Random Under-Sampling Boost[2] (RUSBoost) Classifier (Seiffert et al., 2010). Both algorithms randomly undersample the majority class (in our case, the negative class) to balance the proportion between positive and negative observations.

To select suitable parameters for both classifiers, we applied a randomized parameter search with stratified 10-fold cross-validation. The stratification ensured that the class distributions in each fold remain comparable. In awareness of the risk to overfit parameters to the data (Probst et al., 2019), we performed the parameter search only on the data subset for the *Power buys TV (2016/2017)* prediction case, as it was the subset among the two cases with the largest number of observations that has the higher proportion of true positives (Table 2). We chose to use parameter configurations which resulted in the best performance of the classifiers regarding the AUC metric (we describe the metric in detail in Section 4.1). Thereafter, we trained and evaluated models for each case, year, and ML algorithm with stratified 10-fold cross-validation to mitigate bias in measuring the classifiers' performance.

Beyond the ML models, which are capable to predict, we employed the XAI method SHAP with its implementation for tree-based ML models (TreeSHAP) to illustrate the patterns that ML detected in the data. We integrated our analysis into the stratified 10-fold cross-validation and computed SHAP values for each test fold to observe patterns in the predictions. To determine whether SHAP provides robust explanations, we compare SHAP values between the cross-validation's test folds for customers who purchased an additional product. Further, we conducted statistical tests to investigate if the patterns explained by the XAI method actually hold for the following years.

## 4 Evaluation and Results

We evaluate the predictive performance of the ML models in order to answer RQ1 and we evaluate the outputs of the XAI method SHAP to answer RQ2.

### 4.1 Predictive performance

To assess the quality of the prediction models, we adhere to the standards of ML research and use 10-fold cross-validation with a stratified random sampling (Hastie et al., 2009) for each year, and compare the prediction for each customer with the true customer behavior. For each customer we count the true positive (tp), false positive (fp), true negative (tn), and false negative (fn) predictions. We monitor the performance over time between the different years to account for the longitudinal format of the data. Given the imbalanced class setting, we use several measures of classification performance (Branco et al., 2016): Precision, Recall, the tradeoff between both ($F_\beta$), and the AUC metric, which is "one of the most used measures under imbalanced domains" (Branco et al., 2016, p. 9).

Precision states the ratio of true positives to the group of observations declared as positive by the classifier and computes as *Precision = tp / (tp + fp)*. It takes values between 0 and 1, where 0 implies for our case that the classifier did not correctly predict any cross-purchase in the following year and 1 that all positive predictions were correct.

Recall (also called "sensitivity") quantifies the ability of a classifier to detect observations that are truly positive: *Recall = tp / (tp + fn)*. The range of values for recall is [0,1], where 0 indicates that the classifier did not identify any of the customers who cross-purchase in the following year, and 1 that all customers that made a cross-purchase were recognized by the model.

The $F_\beta$ score is a tradeoff measure calculated from precision and recall that allows to apply the weight $\beta$ to assign a higher value to precision or recall respectively: $F_\beta = (1 + \beta^2) * Precision * Recall / ((\beta^2 * Precision) + Recall)$. Given our case, we selected $\beta = 2$, which weighs recall four times higher than precision. The codomain of $F_\beta$ is [0,1], where 0 indicates that precision or recall are equal to 0 and 1 that both metrics reach their optimum.

---

[1] Balanced RF parameters: *n_estimators*: 1600, *min_samples_split*: 5, *min_samples_leaf*: 2, *max_features*: sqrt, *max_depth*: 50, *bootstrap*: True

[2] RUSBoost parameters: *replacement*: True, *n_estimators*: 200, *learning_rate:* 0.1

Finally, we use the Area Under the Receiver Operating Characteristics Curve (AUC), which indicates whether a classifier can discriminate among the classes. It measures the area below the curve created by plotting the recall against the false positive rate (fp / fp + tn). AUC avoids making assumptions about the tradeoff between precision and recall and is naturally bounded between 0 and 1, whereas 1 indicates a perfect prediction, 0.5 is the expected AUC for a random guess, and 0 a classification which is totally wrong for all the observations (Fawcett, 2006). We illustrate the predictive performance as AUC values in Figure 2 and display the detailed results for the other metrics in Table 3.

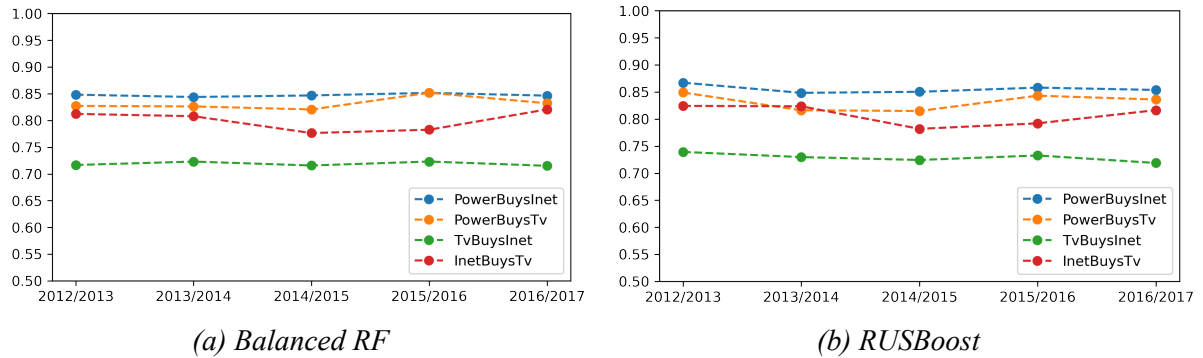

*(a) Balanced RF*            *(b) RUSBoost*

*Figure 2.    Predictive performance (AUC) for the four cross-selling products over five years.*

*We answer our first RQ as follows*: The prediction model for electricity customers signing an Internet contract can correctly predict with up to 86% (AUC) probability whether an individual customer will purchase this product. The predictive performance of the models is satisfactory overall, except for the cross-sell of TV customers to internet contracts. All in all, this result can augment the sales expert's work and can give them confidence, especially when selling a contract that may bind the customer for a long time. In this way, the expert can decide whether a product offer makes sense or whether it would be better to invest the time of the customer conversation into a customer loyalty measure.

| Year (Train/Test) | Balanced RF | | | | RUSBoost | | | |
|---|---|---|---|---|---|---|---|---|
| | AUC | Precision | Recall | F2 | AUC | Precision | Recall | F2 |
| **Power buys Inet** | | | | | | | | |
| 2012/2013 | 0.848 | 0.032 | 0.972 | 0.141 | 0.867 | 0.038 | 0.963 | 0.164 |
| 2013/2014 | 0.844 | 0.035 | 0.969 | 0.153 | 0.848 | 0.038 | 0.948 | 0.165 |
| 2014/2015 | 0.847 | 0.037 | 0.961 | 0.160 | 0.850 | 0.041 | 0.935 | 0.174 |
| 2015/2016 | 0.852 | 0.046 | 0.957 | 0.193 | 0.858 | 0.051 | 0.941 | 0.210 |
| 2016/2017 | 0.846 | 0.042 | 0.940 | 0.178 | 0.854 | 0.047 | 0.920 | 0.197 |
| **Power buys Tv** | | | | | | | | |
| 2012/2013 | 0.827 | 0.043 | 0.836 | 0.177 | 0.849 | 0.047 | 0.870 | 0.193 |
| 2013/2014 | 0.826 | 0.057 | 0.822 | 0.223 | 0.816 | 0.053 | 0.812 | 0.211 |
| 2014/2015 | 0.821 | 0.047 | 0.824 | 0.191 | 0.815 | 0.043 | 0.829 | 0.179 |
| 2015/2016 | 0.852 | 0.106 | 0.833 | 0.351 | 0.843 | 0.098 | 0.826 | 0.332 |
| 2016/2017 | 0.832 | 0.065 | 0.818 | 0.246 | 0.836 | 0.066 | 0.825 | 0.249 |
| **Inet buys Tv** | | | | | | | | |
| 2012/2013 | 0.812 | 0.022 | 0.799 | 0.098 | 0.824 | 0.039 | 0.738 | 0.160 |
| 2013/2014 | 0.808 | 0.029 | 0.775 | 0.126 | 0.824 | 0.051 | 0.730 | 0.200 |
| 2014/2015 | 0.776 | 0.020 | 0.721 | 0.089 | 0.782 | 0.036 | 0.646 | 0.147 |
| 2015/2016 | 0.783 | 0.022 | 0.732 | 0.099 | 0.792 | 0.044 | 0.658 | 0.174 |
| 2016/2017 | 0.821 | 0.053 | 0.730 | 0.205 | 0.816 | 0.095 | 0.677 | 0.303 |
| **Tv buys Inet** | | | | | | | | |
| 2012/2013 | 0.717 | 0.037 | 0.780 | 0.156 | 0.739 | 0.043 | 0.771 | 0.176 |
| 2013/2014 | 0.723 | 0.043 | 0.777 | 0.176 | 0.730 | 0.046 | 0.761 | 0.185 |
| 2014/2015 | 0.716 | 0.038 | 0.781 | 0.158 | 0.724 | 0.041 | 0.758 | 0.169 |
| 2015/2016 | 0.723 | 0.041 | 0.778 | 0.168 | 0.733 | 0.044 | 0.770 | 0.178 |
| 2016/2017 | 0.715 | 0.035 | 0.744 | 0.147 | 0.719 | 0.037 | 0.719 | 0.155 |

*Table 3.    Results of Balanced RF and RUSBoost Classifier.*

## 4.2 SHAP explanation of predictions

SHAP allows to obtain explanations why an ML model predicts a certain outcome. To illustrate the explanation result, we focus on the case "Power buys TV" of the Balanced RF with training data from 2016 to predict cross-purchases in 2017, given that this subset has the higher proportion of true positives among the two cases with the largest number of observations.

Figure 3 displays the SHAP summary plot indicating the estimated influence of the features on the classification result. Each point denotes an observation, with the color representing the level of the feature value on the respective scale (red belongs to higher and blue to lower feature values). The SHAP value determines the position of the points on the x-axis. Hence, the further to the right an observation for the respective feature is displayed on the plot, the stronger the shift of the model decision towards the positive class (here: purchase of a TV contract in 2017). If points are overlapping due to equal SHAP values, they are jittered. Finally, the plot orders the features according to their importance for the model's output on the y-axis.

A sales expert can use these explanations to gain additional insights to tailor sales talks. For example, a larger number of customer relationship months makes it unlikely that an electricity customer also buys a TV contract (feature 7 in Figure 3), or that older customers buy their internet contract from the utility company (feature 9). Yet, the majority of features represent information that is less usable in a sales conversation, like the revenue numbers (e.g., feature 1, 2, and 10) or the geographic location.

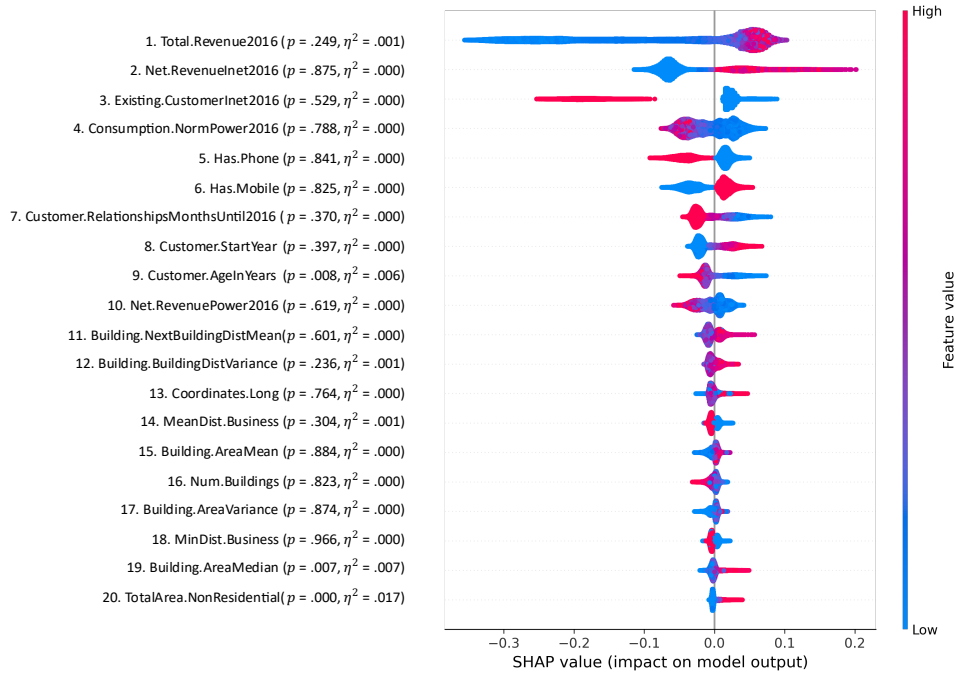

*Figure 3.  Aggregated SHAP summary plot for each test fold of the cross-validation and Kruskal-Wallis test results for the 20 most important feaures.*

To assess the reliability of the explanations given, we perform two analyses. First, we check the robustness based on 10 subsamples of the data and test whether the ML and SHAP approach yield to similar feature attributions in each subsample. This analysis has the underlying assumption that a robust explanation for a feature can be replicated using different data subsets. We integrated our analysis in the stratified 10-fold cross-validation and checked for differences of the SHAP values between the test folds (i.e., groups) for customers that actually cross-purchased (i.e., true positives and false negatives). We assumed explanations to be robust either if the feature attributions assigned show no significant differences for buyers at all, or if they were significant but with a small effect size among the groups. To quantify differences between the groups, we used a non-parametric Kruskal-Wallis (1952) test, given that the SHAP values of the features are not normally distributed. We list the results of this robustness evaluation in the left part of Figure 3 (after the feature name, we display the significance level and the effect size). For users, this information could be encoded, for example, using a traffic light symbol.

The second evaluation concerns the detected patterns and tests if those patterns are also existent in following years. Like a sales expert, we therefore formulated hypotheses based on SHAP-assigned feature attributions. To validate the explanations, we conducted statistical tests using customer data from the following year (in this case 2017) to investigate whether the patterns explained by the XAI method actually hold true and thus can help energy utilities in targeting customers. For the statistical tests, we selected the 10 most influential features for prediction (Figure 3) because there, we could observe a clear pattern. We formulated hypotheses from the patterns (e.g., high values for *Total.Revenue2016* lead customers to sign a TV contract in 2017) and tested them for the two samples (i.e., buyers and non-buyers) by performing a one-sided two-sample Welch's t-test, a one-sided two-sample student's t-test (for the variable *Customer.AgeInYears* due to homogeneous variance between the samples), and a chi-squared test for the binary variables (*Has.Phone* and *Has.Mobile*). We determined the respective effect size Cohens' $d$ and $\omega$ (Cohen, 2013). Table 4 displays the results of these statistical tests.

| #  | Variable                              | df       | Statistic | p      | Eff. Size |
|----|---------------------------------------|----------|-----------|--------|-----------|
| 1  | Total.Revenue2017                     | 1527.73  | 10.96     | < .001 | 0.27      |
| 2  | Net.RevenueInet2017                   | 1518.68  | 22.64     | < .001 | 0.68      |
| 3  | Existing.CustomerInet2017             | 1605.44  | –31.16    | < .001 | –0.42     |
| 4  | Consumption.NormPower2017             | 1521.40  | –1.14     | .128   | –0.03     |
| 5  | Has.Phone                             | 1        | 126.33    | < .001 | 0.03      |
| 6  | Has.Mobile                            | 1        | 297.75    | < .001 | 0.04      |
| 7  | Customer.RelationshipsMonthsUntil2017 | 1518.38  | –14.47    | < .001 | –0.44     |
| 8  | Customer.StartYear                    | 1518.68  | 14.38     | < .001 | 0.43      |
| 9  | Customer.AgeInYears                   | 165092   | –11.23    | < .001 | –0.29     |
| 10 | Net.RevenuePower2017                  | 1520.48  | –0.65     | .257   | –0.02     |

*Table 4. Results of the statistical validation of XAI-detected pattern in the following year.*

*We answer our second RQ as follows*: Given that all characteristics in our evaluation show high robustness, i.e., there are no significant differences, or if significant only with small effect size, we find that the XAI method SHAP can provide reliable explanations. Moreover, we find that the XAI-detected patterns in the data hold for following years and are therefore valid, as there is a significant difference (in the respective direction) between buyers and non-buyers (except for the features *Consumption.NormPower2017* and *Net.RevenuePower2017*). Both findings can increase the confidence of sales agents when using explanations from XAI methods embedded in IS. We thus conclude that XAI methods can overall very satisfactorily support the cross-selling process with detailed insights.

# 5    Discussion and implications

Research and practice has high interest in effective ML business applications. Although we see many successful applications of ML in research studies and single commercial applications, the majority of firms are not exploiting the true potential of ML technology. Only firms in some business areas benefit from their investments in ML and related data analytics technologies (Müller et al., 2018; Tambe, 2014; Wu et al., 2019). One reason for this is the lack of clarity regarding which concrete business applications and processes ML methods can truly complement (Davenport and Ronanki, 2018; Fountaine et al., 2019). Hence, our study explored the use of current ML and XAI methods in the business field of cross-selling in energy retailing.

Operating in a data-rich environment (with CRM systems and data from long customer relations), sales experts struggle to have the right information at hand at the time of interaction. Thus, predictive and prescriptive insights are necessary to augment their hardly automatable work. Having applied ML on a dataset on 220,185 customers from a European utility, we found that this technology can predict which product(s) individual customers buy in the following year. In addition, we demonstrated that XAI methods can give sales experts rich insights on influencing factors why customers might buy these products. These insights pass statistical tests for robustness, and their true existence hold in consecutive years. The results inform two current research areas of ML applications (inscrutability and causality), have implications for relationship marketing research, and lay the foundations for future studies that involve practitioners.

## 5.1 Implications for ML applications

Corporate applications of ML face several practical challenges that research needs to solve (Baier et al., 2019; Lee and Shin, 2020). Based on our empirical case study, we have shown that two of these problems—the inscrutability of models and the lack of causality in explanations generated by XAI—can be overcome with a skillful use of available techniques:

First, many ML applications are *inscrutable* to humans, with negative consequences on the acceptance of ML predictions among users (Burton et al., 2020; Dietvorst et al., 2018). In addition, business users often want prescriptions (e.g., decision recommendations) instead of predictions, or want to exploit the pattern recognition capability of the algorithms (Berente et al., 2019; Tremblay et al., 2021). We explored the use of ML models in that we made a model transparent with a feature attribution XAI approach. We confirmed findings from earlier research (Abdul et al., 2018; Miller, 2019; Wastensteiner et al., 2021) that current results of XAI are not designed for end-user facing applications. Furnishing the data insights with information on the robustness or true existence in other years could increase the acceptance and trust in such insights.

Second, current XAI approaches are criticized to lack *causality* (Rudin, 2019), because variables that are causally related to the business problem can only be identified with the help of human experts and not with automated approaches. At least for our case, where we have data for several years at hand, we could apply statistical tests to demonstrate that the pattern identified by XAI can be confirmed with data of future years.

## 5.2 Implications for relationship marketing theory and practice

Our study extends the scope of ML applications to support relationship marketing processes. Earlier studies focused strongly on targeting issues—which means to select the most promising customers for an offer. We demonstrated that, because of the long history and proliferation of data in relationship marketing, it is also feasible to support the personalization (tailoring) for each individual customer. This means that ML models can provide sales experts with information on which product has the highest likelihood to be purchased and which argument for the product fits the customer. This helps them at the point of customer interaction, where sales experts cannot process huge amounts of information (Sweller, 1993). Targeted insights give them confidence in the sales conversation, avoids wasting valuable contact time, or unprofitable cross-buying behavior (Shah et al., 2012). In this respect, predictions also help to discard existing practices such as hand-prioritized sales lists or gut feelings, and to make data-driven, evidence-based operational decisions. In marketing, customer value can be improved (Kumar, 2018) by increasing the cross-sell probability (offering the right product increases the likelihood of a sale). Customer contact costs can also be reduced, as one can save the futile effort of making offers to a customer who will not buy. Finally, customer satisfaction may be strengthened, as more targeted contact takes place.

## 5.3 Limitations and future work

Our study investigates one of the first applications of XAI in the augmentation of cross-selling agents. Therefore, our study is just at the beginning of a larger research endeavor and comes with some limitations. The most important limitation is that the data we received from the company was, if not indicated with year indices in this paper, static because their systems could not export core data changes over time. Further validations of XAI approaches are therefore necessary.

Another area of improvements is the quantitative modelling. We use variants of the RF and AdaBoost classifiers in our analysis for cross-purchase predictions. We evaluated them with stratified 10-fold cross-validation, and used SHAP for feature attribution explanations. Here, the analysis could be extended by a more extensive parameter search and the use of additional and more sophisticated ML approaches based, for example, on a focal loss function (Lin et al., 2018) and gradient boosted trees (Chen and Guestrin, 2016; Prokhorenkova et al., 2019). Also, the analysis could be extended by a repeated cross-validation, providing a larger amount of data in the folds, which would additionally strengthen the statistical tests performed. In addition, future work could consider other types of XAI explanations, such as counterfactual explanations that provide sales agents with insights on how feature values need to change in order to shift a model's decision to a desired outcome (Mothilal et al., 2020).

Finally, research should quantify the effect of ML predictions and the respective explanations on various performance indicators (i.e., task performance of sales agents, conversion rates, quality of consulting sessions, employee satisfaction, perceived usefulness, trust in predictions, durations of calls, etc.) and moderators (i.e., experienced vs. novice sales agents, good vs. bad performing sales agents). For this, controlled field experiments should be conducted. One central question is also, if the predictions work alone or only in combination with explanations. A follow-up question that can be part of this field validation is how XAI methods perform compared to established descriptive approaches (e.g., from visualizations or business intelligence systems) in terms of leveraging existing patterns from the data. In addition, how visualizations should be designed in order to gain insights from XAI in the field.

# 6    Conclusion

Our study has confirmed that ML and XAI methods offer great opportunities to support sales experts in cross-selling activities. In the presented case, these methods could transform comprehensive data from customer relations into actionable insights (predictions regarding the purchase probability of several products for individual customers and corresponding explanations on influencing factors). Furthermore, our study creates a basis to examine this application in future field studies and to explore the mechanisms that lead to ML and XAI being successfully used in organizations.

## Acknowledgements

We would like to thank the partnering companies for access to the data. Our personal thanks go in particular to Tobias Graml (BEN Energy, Zurich / Munich) for his feedback on our analysis. The research presented in this paper was financially supported by the EUREKA member countries and the European Union (Eurostars grant number E!114466 - BENEFIZZO).

## References

Abdul, A., Vermeulen, J., Wang, D., Lim, B.Y., Kankanhalli, M., 2018. Trends and Trajectories for Explainable, Accountable and Intelligible Systems: An HCI Research Agenda, in: Proceedings of the 2018 CHI Conference on Human Factors in Computing Systems, CHI '18. Association for Computing Machinery, New York, NY, USA, pp. 1–18. https://doi.org/10.1145/3173574.3174156

Adadi, A., Berrada, M., 2018. Peeking Inside the Black-Box: A Survey on Explainable Artificial Intelligence (XAI). IEEE Access 6, 52138–52160. https://doi.org/10.1109/ACCESS.2018.2870052

Akter, S., Fosso Wamba, S., 2016. Big data analytics in E-commerce: a systematic review and agenda for future research. Electron Markets 26, 173–194. https://doi.org/10.1007/s12525-016-0219-0

Albert, T., Goes, P., 2004. GIST: A Model for Design and Management of Content and Interactivity of Customer-Centric Web Sites. MIS Quarterly 28.

Arnott, D., Pervan, G., 2008. Eight key issues for the decision support systems discipline. Decision Support Systems 44, 657–672. https://doi.org/10.1016/j.dss.2007.09.003

Baier, L., Jöhren, F., Seebacher, S., 2019. Challenges in the Deployment and Operation of Machine Learning in Practice, in: ECIS 2019 Proceedings. AIS electronic library.

Bailey, D.E., Barley, S.R., 2020. Beyond design and use: How scholars should study intelligent technologies. Information and Organization 30, 100286. https://doi.org/10.1016/j.infoandorg.2019.100286

Barredo Arrieta, A., Díaz-Rodríguez, N., Del Ser, J., Bennetot, A., Tabik, S., Barbado, A., Garcia, S., Gil-Lopez, S., Molina, D., Benjamins, R., Chatila, R., Herrera, F., 2020. Explainable Artificial Intelligence (XAI): Concepts, taxonomies, opportunities and challenges toward responsible AI. Information Fusion 58, 82–115. https://doi.org/10.1016/j.inffus.2019.12.012

Berente, N., Gu, B., Recker, J., Santhanam, R., 2021. Managing Artificial Intelligence. MIS Quarterly 45, 1433–1450.

Berente, N., Seidel, S., Safadi, H., 2019. Research Commentary—Data-Driven Computationally Intensive Theory Development. Information Systems Research 30, 50–64. https://doi.org/10.1287/isre.2018.0774

Binbasioglu, M., Jarke, M., 1986. Domain specific DSS tools for knowledge-based model building. Decision Support Systems 2, 213–223. https://doi.org/10.1016/0167-9236(86)90029-1

Branco, P., Torgo, L., Ribeiro, R.P., 2016. A Survey of Predictive Modeling on Imbalanced Domains. ACM Comput. Surv. 49, 31:1-31:50. https://doi.org/10.1145/2907070

Brynjolfsson, E., Rock, D., Syverson, C., 2017. Artificial Intelligence and the Modern Productivity Paradox: A Clash of Expectations and Statistics (NBER Working Paper 24001 No. w24001). National Bureau of Economic Research, Cambridge, MA. https://doi.org/10.3386/w24001

Burton, J.W., Stein, M.-K., Jensen, T.B., 2020. A systematic review of algorithm aversion in augmented decision making. Journal of Behavioral Decision Making 33, 220–239. https://doi.org/10.1002/bdm.2155

Carvalho, D.V., Pereira, E.M., Cardoso, J.S., 2019. Machine learning interpretability: A survey on methods and metrics. Electronics 8, 832.

Chang, H.H., Wong, K.H., Fang, P.W., 2014. The effects of customer relationship management relational information processes on customer-based performance. Decision Support Systems 66, 146–159. https://doi.org/10.1016/j.dss.2014.06.010

Chen, C., Liaw, A., Breiman, L., 2004. Using Random Forest to Learn Imbalanced Data. URL https://statistics.berkeley.edu/sites/default/files/tech-reports/666.pdf (accessed 11.12.21).

Chen, T., Guestrin, C., 2016. XGBoost: A Scalable Tree Boosting System, in: Proceedings of the 22nd ACM SIGKDD International Conference on Knowledge Discovery and Data Mining. Presented at the KDD '16: The 22nd ACM SIGKDD International Conference on Knowledge Discovery and Data Mining, ACM, San Francisco, California, USA, pp. 785–794. https://doi.org/10.1145/2939672.2939785

Cheng, H.-F., Wang, R., Zhang, Z., O'Connell, F., Gray, T., Harper, F.M., Zhu, H., 2019. Explaining Decision-Making Algorithms through UI: Strategies to Help Non-Expert Stakeholders, in: Proceedings of the 2019 CHI Conference on Human Factors in Computing Systems - CHI '19. Presented at the the 2019 CHI Conference, ACM Press, Glasgow, Scotland Uk, pp. 1–12. https://doi.org/10.1145/3290605.3300789

Cohen, J., 2013. Statistical Power Analysis for the Behavioral Sciences, 0 ed. Routledge. https://doi.org/10.4324/9780203771587

Coltman, T., 2007. Why build a customer relationship management capability? The Journal of Strategic Information Systems 16, 301–320. https://doi.org/10.1016/j.jsis.2007.05.001

Coltman, T., Devinney, T.M., Midgley, D.F., 2011. Customer relationship management and firm performance. J Inf Technol 26, 205–219. https://doi.org/10.1057/jit.2010.39

Constantiou, I.D., Kallinikos, J., 2015. New games, new rules: big data and the changing context of strategy. J Inf technol 30, 44–57. https://doi.org/10.1057/jit.2014.17

Cramton, P., 2017. Electricity market design. Oxford Review of Economic Policy 33, 589–612. https://doi.org/10.1093/oxrep/grx041

Cui, G., Wong, M.L., Wan, X., 2012. Cost-Sensitive Learning via Priority Sampling to Improve the Return on Marketing and CRM Investment. Journal of Management Information Systems 29, 341–374.

Cullen, J., Bryman, A., 1988. The Knowledge Acquisition Bottleneck: Time for Reassessment? Expert Systems 5, 216–225. https://doi.org/10.1111/j.1468-0394.1988.tb00065.x

Davenport, T.H., Ronanki, R., 2018. Artificial Intelligence for the Real World. Harvard Business Review.

Dietvorst, B.J., Simmons, J.P., Massey, C., 2018. Overcoming Algorithm Aversion: People Will Use Imperfect Algorithms If They Can (Even Slightly) Modify Them. Management Science 64, 1155–1170. https://doi.org/10.1287/mnsc.2016.2643

Duan, Y., Edwards, J.S., Dwivedi, Y.K., 2019. Artificial intelligence for decision making in the era of Big Data – evolution, challenges and research agenda. International Journal of Information Management 48, 63–71. https://doi.org/10.1016/j.ijinfomgt.2019.01.021

Emanet, N., Öz, H.R., Bayram, N., Delen, D., 2014. A comparative analysis of machine learning methods for classification type decision problems in healthcare. Decis. Anal. 1, 6. https://doi.org/10.1186/2193-8636-1-6

Eurostat, 2017. Final consumption expenditure of households, by consumption purpose - Eurostat (Code: tsdpc520, Last update: 25/01/17 ) [WWW Document]. URL http://ec.europa.eu/eurostat/web/products-datasets/-/tsdpc520 (accessed 1.25.17).

Fawcett, T., 2006. An introduction to ROC analysis. Pattern Recognition Letters 27, 861–874.

Fayyad, U., Piatetsky-Shapiro, G., Smyth, P., 1996. The KDD Process for Extracting Useful Knowledge from Volumes of Data. Commun. ACM 39, 27–34. https://doi.org/10.1145/240455.240464

Fernandez-Delgado, M., Cernadas, E., Barro, S., Amorim, D., 2014. Do we Need Hundreds of Classifiers to Solve Real World Classification Problems? Journal of Machine Learning Research 15, 3133–3181.

Fountaine, T., McCarthy, B., Saleh, T., 2019. Building the AI-Powered Organization. Harvard Business Review.

Gefen, D., Ridings, C.M., 2002. Implementation Team Responsiveness and User Evaluation of Customer Relationship Management: A Quasi-Experimental Design Study of Social Exchange Theory. Journal of Management Information Systems 19, 47–69.

Gregor, S., Benbasat, I., 1999. Explanations from intelligent systems: Theoretical foundations and implications for practice. MIS Quarterly 497–530.

Grønsund, T., Aanestad, M., 2020. Augmenting the algorithm: Emerging human-in-the-loop work configurations. The Journal of Strategic Information Systems 101614. https://doi.org/10.1016/j.jsis.2020.101614

Guidotti, R., Monreale, A., Ruggieri, S., Turini, F., Giannotti, F., Pedreschi, D., 2018. A Survey of Methods for Explaining Black Box Models. ACM Comput. Surv. 51, 93:1-93:42. https://doi.org/10.1145/3236009

Haenlein, M., Kaplan, A., 2019. A Brief History of Artificial Intelligence: On the Past, Present, and Future of Artificial Intelligence. California Management Review 61, 5–14. https://doi.org/10.1177/0008125619864925

Hastie, T., Tibshirani, R., Friedman, J., 2009. The Elements of Statistical Learning, Springer Series in Statistics. Springer New York. https://doi.org/10.1007/978-0-387-84858-7

Hopf, K., 2018. Mining Volunteered Geographic Information for Predictive Energy Data Analytics. Energy Informatics. https://doi.org/10.1186/s42162-018-0009-3

Hopf, K., Sodenkamp, M., Kozlovskiy, I., 2016. Energy data analytics for improved residential service quality and energy efficiency, in: ECIS 2016 Research in Progress Proceedings. Presented at the 24. European Conference on Information Systems (ECIS), AIS electronic library, Istanbul, Turkey.

Hsieh, J.J.P.-A., Rai, A., Petter, S., Zhang, T., 2012. Impact of User Satisfaction with Mandated CRM Use on Employee Service Quality. MIS Quarterly 36, 1065–1080.

Kamakura, W.A., 2008. Cross-selling: Offering the right product to the right customer at the right time. Journal of Relationship Marketing 6, 41–58.

Karimi, J., Somers, T.M., Gupta, Y.P., 2001. Impact of Information Technology Management Practices on Customer Service. Journal of Management Information Systems 17, 125–158.

Kim, S.H., Mukhopadhyay, T., 2010. Determining Optimal CRM Implementation Strategies. Information Systems Research 22, 624–639. https://doi.org/10.1287/isre.1100.0309

King, W.R., 1978. Strategic Planning for Management Information Systems. MIS Quarterly 2, 27–37. https://doi.org/10.2307/249104

Kitchens, B., Dobolyi, D., Li, J., Abbasi, A., 2018. Advanced Customer Analytics: Strategic Value Through Integration of Relationship-Oriented Big Data. Journal of Management Information Systems 35, 540–574. https://doi.org/10.1080/07421222.2018.1451957

Kotler, P., Armstrong, G., Harris, L.C., 2017. Principles of marketing, Seventh European Edition. ed. Pearson, Harlow, United Kingdom.

Kruskal, W.H., Wallis, W.A., 1952. Use of ranks in one-criterion variance analysis. Journal of the American Statistical Association 47, 583–621. https://doi.org/10.1080/01621459.1952.10483441

Kühl, N., Hirt, R., Baier, L., Schmitz, B., Satzger, G., 2021. How to Conduct Rigorous Supervised Machine Learning in Information Systems Research: The Supervised Machine Learning Reportcard. Communications of the Association for Information Systems.

Kumar, V., 2018. A Theory of Customer Valuation: Concepts, Metrics, Strategy, and Implementation. Journal of Marketing 82, 1–19. https://doi.org/10.1509/jm.17.0208

LaValle, S., Lesser, E., Hopkins, M.S., Kruschwitz, N., 2011. Big Data, Analytics and the Path From Insights to Value. MIT Sloan Management Review 52.

Lee, I., Shin, Y.J., 2020. Machine learning for enterprises: Applications, algorithm selection, and challenges. Business Horizons, ARTIFICIAL INTELLIGENCE AND MACHINE LEARNING 63, 157–170. https://doi.org/10.1016/j.bushor.2019.10.005

Lee, J.-N., Pi, S.-M., Kwok, R.C., Huynh, M.Q., 2003. The Contribution of Commitment Value in Internet Commerce: An Empirical Investigation. Journal of the Association for Information Systems 4.

Li, M., Gregor, S., 2011. Outcomes of effective explanations: Empowering citizens through online advice. Decision Support Systems 52, 119–132. https://doi.org/10.1016/j.dss.2011.06.001

Lin, T.-Y., Goyal, P., Girshick, R., He, K., Dollár, P., 2018. Focal Loss for Dense Object Detection. arXiv:1708.02002 [cs].

Loureiro, A.L.D., Miguéis, V.L., da Silva, L.F.M., 2018. Exploring the use of deep neural networks for sales forecasting in fashion retail. Decision Support Systems 114, 81–93. https://doi.org/10.1016/j.dss.2018.08.010

Luftman, J., Zadeh, H.S., Derksen, B., Santana, M., Rigoni, E.H., Huang, Z. (David), 2012. Key information technology and management issues 2011–2012: an international study. J Inf Technol 27, 198–212. https://doi.org/10.1057/jit.2012.14

Lundberg, S.M., Erion, G., Chen, H., DeGrave, A., Prutkin, J.M., Nair, B., Katz, R., Himmelfarb, J., Bansal, N., Lee, S.-I., 2020. From local explanations to global understanding with explainable AI for trees. Nature Machine Intelligence 2, 56–67. https://doi.org/10.1038/s42256-019-0138-9

Lundberg, S.M., Lee, S.-I., 2017. A Unified Approach to Interpreting Model Predictions, in: Advances in Neural Information Processing Systems. Curran Associates, Inc., pp. 4765–4774.

Martens, D., Provost, F., Clark, J., de Fortuny, E.J., 2016. Mining Massive Fine-Grained Behavior Data to Improve Predictive Analytics. MIS Quarterly 40, 869–888.

Martínez, A., Schmuck, C., Pereverzyev, S., Pirker, C., Haltmeier, M., 2020. A machine learning framework for customer purchase prediction in the non-contractual setting. European Journal of Operational Research 281, 588–596. https://doi.org/10.1016/j.ejor.2018.04.034

Mehdiyev, N., Fettke, P., 2020. Prescriptive Process Analytics with Deep Learning and Explainable Artificial Intelligence, in: ECIS 2020 Research Papers. Presented at the 28. European Conference on Information Systems, AIS electronic library, Virual, p. 18.

Miller, T., 2019. Explanation in artificial intelligence: Insights from the social sciences. Artificial Intelligence 267, 1–38. https://doi.org/10.1016/j.artint.2018.07.007

Molnar, C., 2019. Interpretable machine learning - A Guide for Making Black Box Models Explainable.

Mothilal, R.K., Sharma, A., Tan, C., 2020. Explaining machine learning classifiers through diverse counterfactual explanations, in: Proceedings of the 2020 Conference on Fairness, Accountability, and Transparency. Presented at the FAT* '20: Conference on Fairness, Accountability, and Transparency, ACM, Barcelona Spain, pp. 607–617. https://doi.org/10.1145/3351095.3372850

Müller, O., Fay, M., Brocke, J. vom, 2018. The Effect of Big Data and Analytics on Firm Performance: An Econometric Analysis Considering Industry Characteristics. Journal of Management Information Systems 35, 488–509. https://doi.org/10.1080/07421222.2018.1451955

Olson, D.L., Chae, B., 2012. Direct marketing decision support through predictive customer response modeling. Decision Support Systems 54, 443–451. https://doi.org/10.1016/j.dss.2012.06.005

Patton, M.Q., 2002. Qualitative research & evaluation methods, 3. ed. ed. Sage, Thousand Oaks, CA.

Probst, P., Wright, M.N., Boulesteix, A.-L., 2019. Hyperparameters and tuning strategies for random forest. WIREs Data Mining Knowl Discov 9. https://doi.org/10.1002/widm.1301

Prokhorenkova, L., Gusev, G., Vorobev, A., Dorogush, A.V., Gulin, A., 2019. CatBoost: unbiased boosting with categorical features. arXiv:1706.09516 [cs].

Ribeiro, M.T., Singh, S., Guestrin, C., 2016. "Why Should I Trust You?": Explaining the Predictions of Any Classifier, in: Proceedings of the 22nd ACM SIGKDD International Conference on Knowledge Discovery and Data Mining, KDD '16. Association for Computing Machinery, New York, NY, USA, pp. 1135–1144. https://doi.org/10.1145/2939672.2939778

Rudin, C., 2019. Stop explaining black box machine learning models for high stakes decisions and use interpretable models instead. Nature Machine Intelligence 1, 206–215. https://doi.org/10.1038/s42256-019-0048-x

Schlegel, U., Arnout, H., El-Assady, M., Oelke, D., Keim, D.A., 2019. Towards A Rigorous Evaluation Of XAI Methods On Time Series, in: 2019 IEEE/CVF International Conference on Computer Vision Workshop (ICCVW). Presented at the 2019 IEEE/CVF International Conference on Computer Vision Workshop (ICCVW), IEEE, Seoul, South Korea, pp. 4197–4201. https://doi.org/10.1109/ICCVW.2019.00516

Schmitz, C., You-Cheong Lee, Lilien, G.L., 2014. Cross-Selling Performance in Complex Selling Contexts: An Examination of Supervisory- and Compensation-Based Controls. Journal of Marketing 78, 1–19.

Seiffert, C., Khoshgoftaar, T.M., Van Hulse, J., Napolitano, A., 2010. RUSBoost: A Hybrid Approach to Alleviating Class Imbalance. IEEE Trans. Syst., Man, Cybern. A 40, 185–197. https://doi.org/10.1109/TSMCA.2009.2029559

Shah, D., Kumar, V., Qu, Y., Chen, S., 2012. Unprofitable Cross-Buying: Evidence from Consumer and Business Markets. Journal of Marketing 76, 78–95. https://doi.org/10.1509/jm.10.0445

Shaw, M.J., Tu, P.-L., 1988. Applying Machine Learning to Model Management in Decision Support Systems ~g. Decision Support Systems 285–305.

Shearer, C., 2000. The CRISP-DM model: the new blueprint for data mining. Journal of data warehousing 5, 13–22.

Shollo, A., Galliers, R.D., 2016. Towards an understanding of the role of business intelligence systems in organisational knowing. Information Systems Journal 26, 339–367. https://doi.org/10.1111/isj.12071

Shrivastava, U., Jank, W., 2015. A data driven framework for early prediction of customer response to promotions, in: AMCIS 2015 Proceedings. Presented at the Americas Conference on Information Systems (AMCIS), AIS electronic library, Puerto Rico.

Sivarajah, U., Kamal, M.M., Irani, Z., Weerakkody, V., 2017. Critical analysis of Big Data challenges and analytical methods. Journal of Business Research 70, 263–286. https://doi.org/10.1016/j.jbusres.2016.08.001

Slack, D., Hilgard, S., Jia, E., Singh, S., Lakkaraju, H., 2020. Fooling LIME and SHAP: Adversarial Attacks on Post hoc Explanation Methods, in: Proceedings of the AAAI/ACM Conference on AI, Ethics, and Society. Presented at the AIES '20: AAAI/ACM Conference on AI, Ethics, and Society, ACM, New York NY USA, pp. 180–186. https://doi.org/10.1145/3375627.3375830

Stiglic, G., Kocbek, P., Fijacko, N., Zitnik, M., Verbert, K., Cilar, L., 2020. Interpretability of machine learning-based prediction models in healthcare. WIREs Data Mining Knowl Discov 10. https://doi.org/10.1002/widm.1379

Sweller, J., 1993. Some cognitive processes and their consequences for the organisation and presentation of information. Australian Journal of Psychology 45, 1–8. https://doi.org/10.1080/00049539308259112

Syam, N., Sharma, A., 2018. Waiting for a sales renaissance in the fourth industrial revolution: Machine learning and artificial intelligence in sales research and practice. Industrial Marketing Management 69, 135–146. https://doi.org/10.1016/j.indmarman.2017.12.019

Tambe, P., 2014. Big Data Investment, Skills, and Firm Value. Management Science 60, 1452–1469. https://doi.org/10.1287/mnsc.2014.1899

Tarafdar, M., Beath, C.M., Ross, J.W., 2019. Using AI to Enhance Business Operations. MIT Sloan Management Review 37–44.

Tremblay, M.C., Kohli, R., Forsgren, N., 2021. Theories in flux: Reimagining theory building in the age of machine learning. MIS Quarterly 45, 455–459.

van den Broek, E., Sergeeva, A., Huysman, M., 2021. When the machine meets the expert: An ethnography of developing AI for hiring. MIS Quarterly.

Velmurugan, M., Ouyang, C., Moreira, C., Sindhgatta, R., 2021. Evaluating Stability of Post-hoc Explanations for Business Process Predictions, in: Hacid, H., Kao, O., Mecella, M., Moha, N., Paik, H. (Eds.), Service-Oriented Computing, Lecture Notes in Computer Science. Springer International Publishing, Cham, pp. 49–64. https://doi.org/10.1007/978-3-030-91431-8_4

Wang, W., Benbasat, I., 2016. Empirical assessment of alternative designs for enhancing different types of trusting beliefs in online recommendation agents. Journal of Management Information Systems 33, 744–775. https://doi.org/10.1080/07421222.2016.1243949

Wang, W., Benbasat, I., 2007. Recommendation agents for electronic commerce: Effects of explanation facilities on trusting beliefs. Journal of Management Information Systems 23, 217–246. https://doi.org/10.2753/MIS0742-1222230410

Wang, W., Wang, M., 2019. Effects of Sponsorship Disclosure on Perceived Integrity of Biased Recommendation Agents: Psychological Contract Violation and Knowledge-Based Trust Perspectives. Information Systems Research 30, 507–522. https://doi.org/10.1287/isre.2018.0811

Ward, J., Hemingway, C., Daniel, E., 2005. A framework for addressing the organisational issues of enterprise systems implementation. The Journal of Strategic Information Systems, Understanding the Contextual Influences on Enterprise Systems (Part II) 14, 97–119. https://doi.org/10.1016/j.jsis.2005.04.005

Wastensteiner, J., Weiss, T.M., Haag, F., Hopf, K., 2021. Explainable AI for tailored electricity consumption feedback–An experimental evaluation of visualizations, in: ECIS 2021 Research Papers. Presented at the 29. European Conference on Information Systems (ECIS), AIS electronic library, Marrakesh, Marrocco.

Watson, H., 2017. Preparing for the Cognitive Generation of Decision Support. MIS Quarterly Executive 16.

Wu, L., Hitt, L., Lou, B., 2019. Data Analytics, Innovation, and Firm Productivity. Management Science 66, 2017–2039. https://doi.org/10.1287/mnsc.2018.3281

Yeomans, M., Shah, A., Mullainathan, S., Kleinberg, J., 2019. Making sense of recommendations. Journal of Behavioral Decision Making 32, 403–414. https://doi.org/10.1002/bdm.2118

Yin, R.K., 2018. Case study research and applications, Sixth Edition. ed. Sage, Los Angeles ; London ; New Delhi ; Singapore ; Washington DC ; Melbourne.

Zablah, A.R., Bellenger, D.N., Straub, D.W., Johnston, W.J., 2012. Performance Implications of CRM Technology Use: A Multilevel Field Study of Business Customers and Their Providers in the Telecommunications Industry. Information Systems Research 23, 418–435. https://doi.org/10.1287/isre.1120.0419